\documentclass{article}
\usepackage{spconf,amsmath,amssymb,graphicx,booktabs,multirow,xcolor}
\usepackage{microtype,xspace}
\usepackage[hidelinks]{hyperref}
\newcommand{\method}{AlbumVSR\xspace}
\newcommand{\R}{\mathbb{R}}
\newcommand{\best}[1]{\textbf{#1}}
\newcommand{\second}[1]{\underline{#1}}

\title{ALBUMVSR: IDENTITY-PRESERVING FACE VIDEO SUPER-RESOLUTION\\
FROM A PERSONAL PHOTO ALBUM}
\name{
Nianyao Du$^{*}$ \qquad
Xijun Wang\sthanks{These authors contributed equally to this work.}
\qquad Stanley H. Chan
}
\address{Purdue University, West Lafayette, Indiana, USA}

\begin{document}
\ninept
\maketitle

\begin{abstract}
Severe blur, downsampling, noise, and compression can erase the identity cues needed to faithfully restore a person's face. We therefore use photographs of the same person as an external source of identity. \method conditions a frozen, one-step streaming diffusion transformer on a personal album. We use four MagFace/CLIP set tokens to encode the album, and 64 texture tokens which come from a fused UV atlas rendered at the current window's estimated face pose. These tokens are appended to the transformer's cached cross-attention context. On an identity-disjoint VoxCeleb2 benchmark with heavy synthetic degradation, \method raises ArcFace similarity to the ground truth from 0.586 to 0.670 and face PSNR from 24.98 to 26.71 dB. On the 91 test identities, its mean ID-GT is higher than that of all 14 baselines.

\end{abstract}


\section{Introduction}
\label{sec:intro}
Severe degradation can erase the facial evidence needed to preserve identity during generative video restoration. A personal album supplies evidence from other views and sessions, but existing personalized restorers largely process individual face images~\cite{refldm,instantrestore,faceme}. Reference-guided video methods may use iterative generation~\cite{rgfvr} or per-identity fine-tuning~\cite{showpolish}.

We condition the frozen, one-step streaming FlashVSR-tiny backbone~\cite{flashvsr} on a photo album. Four permutation-invariant set tokens encode identity, while 64 tokens from a view-aligned UV render provide texture context for each window. Both token sets augment the cached cross-attention context through frozen projections, while LoRA adapters~\cite{lora} adapt the backbone.

We evaluate AlbumVSR on an identity-disjoint VoxCeleb2 benchmark with session-disjoint albums and light and heavy degradation. Across 91 test identities, we compare \method with 14 baselines using identity, fidelity, temporal, and no-reference quality metrics.

\vspace{-2mm}
\section{Related Work}
\label{sec:related}

\textbf{Video and face restoration.}
One-step diffusion lowers the cost of generative VSR. Flash\-VSR~\cite{flashvsr} and SeedVR2~\cite{seedvr2} reduce sampling to a single step, and FlashVSR further carries attention state across windows. We use FlashVSR as the frozen backbone. RealViFormer~\cite{realviformer} instead targets real-world VSR with channel attention. Face-specific video restorers add stronger facial or temporal priors: KEEP~\cite{keep} propagates a Kalman-inspired face prior, PGTFormer~\cite{pgtformer} predicts temporally coherent parsing-guided codes, SVFR~\cite{svfr} builds on Stable Video Diffusion, and DicFace~\cite{dicface} couples a Dirichlet-constrained variational codebook with temporal coherence. These methods do not use an external photograph of the target
person at inference.

\textbf{Personalized restoration.}
Reference photos can supply identity details missing from degraded inputs. ReF-LDM~\cite{refldm} caches keys and values from a flexible reference set, InstantRestore~\cite{instantrestore} uses shared-image attention to restore a face image from one to four references in a single forward pass, and FaceMe~\cite{faceme} conditions blind face restoration on personal images. For video, Show and Polish (IP-FVR) fine-tunes LoRA weights for each identity and uses feedback learning and cross-clip blending to limit identity drift~\cite{showpolish}. RGFVR~\cite{rgfvr} uses a single reference in a flow-matching video model. 

\begin{figure*}[t]
\centering
\includegraphics[width=1.0\textwidth]{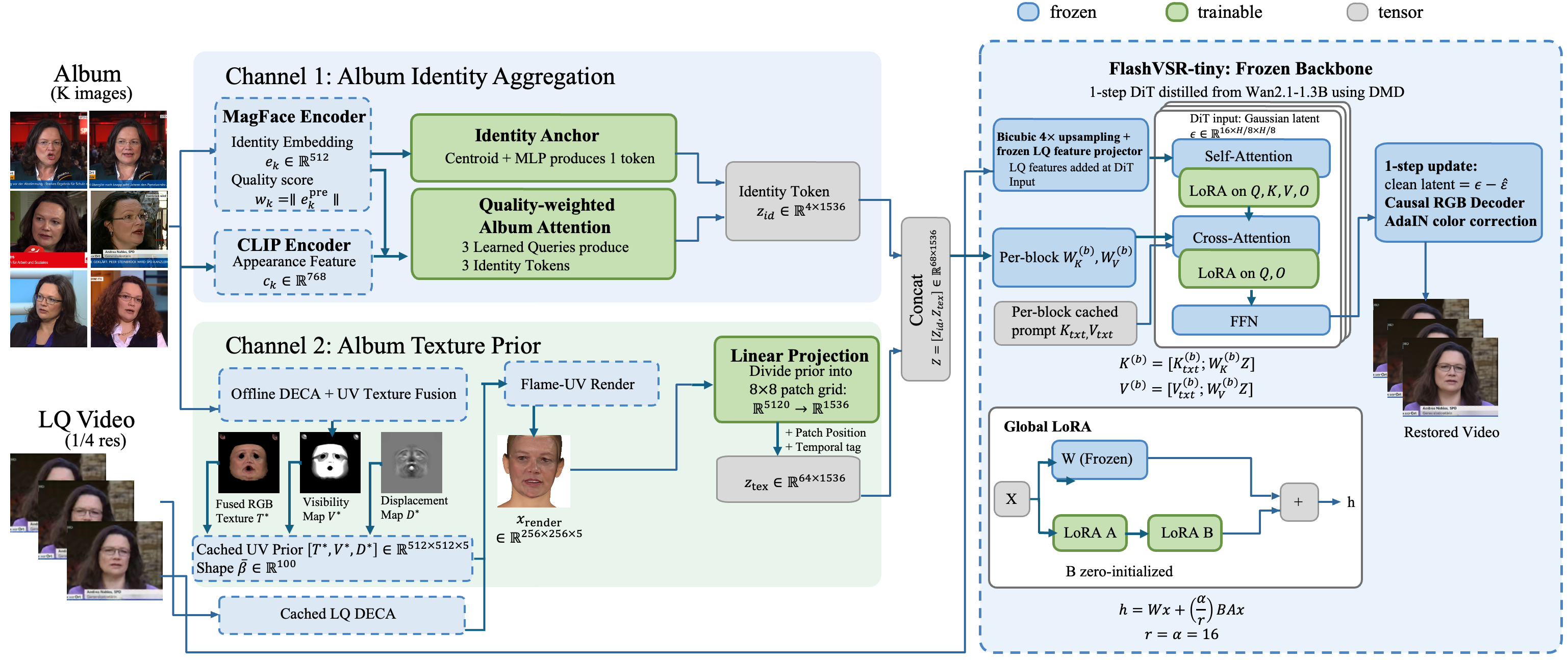}
\caption{Overview of \method. The album is encoded once for each identity. Four set tokens carry identity information, while fused texture, visibility, and displacement maps are rendered and tokenized for each streaming window. Every frozen DiT block appends the resulting album K/V to its cached prompt K/V. Training updates only the album heads and LoRA parameters.}
\label{fig:pipeline}
\end{figure*}

\vspace{-2mm}
\section{Method}
\label{sec:method}

\vspace{-1mm}
\subsection{Album identity tokens}
For each album image $I_k$, frozen MagFace~\cite{magface} and CLIP ~\cite{clip} encoders produce a unit identity embedding $e_k\in\R^{512}$, MagFace quality $w_k=\|e_k^{\rm pre}\|$, and appearance descriptor $c_k\in\R^{768}$. We compute a provisional quality-weighted center, retain references whose cosine similarity to that center is at least $\tau=0.5$, and recompute the center over the retained set:
\begin{equation}
\begin{aligned}
\bar e^{(0)}
&= \operatorname{norm}\!\left(\sum_k w_k e_k\right),
&\quad
\mathcal K
&= \left\{k : e_k^\top \bar e^{(0)} \ge \tau\right\},\\
\bar e
&= \operatorname{norm}\!\left(\sum_{k\in\mathcal K} w_k e_k\right).
\end{aligned}
\label{eq:anchor}
\end{equation}
If this rejection would remove every reference, we keep the highest-quality valid image. An MLP maps $\bar e$ to an anchor token. To retain information beyond the pooled center, we project $[e_k;c_k]$ into a 512-D memory and read it with three learned queries. Their attention logits include $\log w_k$, with invalid and rejected images masked. Together, the anchor and three query outputs form $Z_{\rm id}\in\R^{4\times1536}$, which we call Channel 1 (Ch.1). This permutation-invariant set operation accepts any album size $K$ and reduces naturally to a single reference when $K=1$, following~\cite{nan}.

\vspace{-1mm}
\subsection{Album texture tokens}
We fuse accepted album views in the FLAME UV domain~\cite{flame}, masking occluded and nonfacial pixels~\cite{bisenet}. Quality-weighted averaging supplies low-frequency texture. At each location, the highest-weight view supplies Laplacian detail to avoid blur from misregistered fits. The offline output comprises a $512^2$ RGB atlas $T^\star$, visibility map $V^\star$, displacement map $D^\star$, and pooled face shape.

For each streaming window, cached LQ-video DECA~\cite{deca} parameters pose the pooled mesh at a face-valid frame near the window center. We rasterize the posed mesh with nvdiffrast~\cite{nvdiffrast} and sample $[T^\star,V^\star,D^\star]$ to obtain a $256^2$ five-channel view. An $8\times8$ patch grid is projected to 64 tokens in $\mathbb{R}^{1536}$, with learned positions and a temporal tag. We call the identity and texture paths Ch.1 and Ch.2.

\vspace{-1mm}
\subsection{Streaming adaptation and objective}

FlashVSR-tiny combines a DMD-distilled~\cite{dmd} Wan2.1-1.3B DiT~\cite{wan} with an LQ projection stream, cached self-attention, and a small causal decoder. It processes 21 frames in the first window and advances by eight frames thereafter. Album conditioning must therefore change from one window to the next without invalidating the static prompt cache. Let $K_{\rm txt},V_{\rm txt}$ denote that frozen cache and $Z=[Z_{\rm id};Z_{\rm tex}]$. Each cross-attention block forms

\begin{equation}
 K=[K_{\rm txt};W_KZ],\qquad V=[V_{\rm txt};W_VZ],
 \label{eq:kv}
\end{equation}

where $W_K,W_V$ are frozen but preserve gradients to the album heads. We add LoRA to self-attention $Q,K,V,O$ and cross-attention $Q,O$:

\begin{equation}
 h=Wx+\frac{\alpha}{r}BAx,\qquad r=\alpha=16,
 \label{eq:lora}
\end{equation}

with $B$ initialized to zero. These updates act globally over the latent grid. We leave cross-attention $K$ and $V$ unchanged because adapting them would invalidate the cached prompt K/V.

Following FlashVSR, decoded frames receive AdaIN colour correction against the bicubic input during both training and inference.

The loss weights are 1.0 for spatially weighted global L1, 0.2 for VGG, 1.0 for face L1, 0.25 for face Laplacian, 0.5 for face VGG-LPIPS, 0.2 for CLIP temporal drift, and 0.5 for MagFace identity. The global L1 uses face/background weights of 1.0/0.2. The identity term combines cosine distance to the matching HQ frame and album center $\bar e$ with weights 0.8/0.2. VGG, LPIPS, and identity use four frames per clip (VGG at half resolution), and drift uses six. Evaluation instead uses ArcFace~\cite{arcface} and AlexNet LPIPS~\cite{lpips}.

\begin{table*}[t]
    \centering
    \caption{Heavy-degradation results on 91 test identities. $\dagger$: crop-based method under the shared paste-back protocol. SSIM-F is face-region SSIM. Runtime devices are A100 80GB PCIe ($A$) and RTX 4070 Ti ($W$). VIDD is reported descriptively rather than ranked (Sec.~\ref{sec:experiments}). Best and second-best values in the remaining metrics are bold and underlined.}
    \label{tab:main}
    \setlength{\tabcolsep}{1.8pt}
    \renewcommand{\arraystretch}{0.94}
    \begin{tabular*}{0.99\textwidth}{@{\extracolsep{\fill}}lrrrrrrrrr}
    \toprule
    Method & ID-GT$\uparrow$ & ID-alb$\uparrow$ & PSNR-F$\uparrow$ & SSIM-F$\uparrow$ & LPIPS-F$\downarrow$ & Warp$\downarrow$ & VIDD & MUSIQ$\uparrow$ & s/fr$\downarrow$\\
    \midrule 
    Bicubic $4\times$ & 0.611 & 0.474 & 25.24 & 0.758 & 0.379 & \second{0.0059} & 0.056 & 19.80 & -\\
    FlashVSR-tiny (frozen)~\cite{flashvsr} & 0.586 & 0.459 & 24.98 & 0.704 & 0.175 & 0.0093 & 0.049 & \second{72.34} & 0.07$^{W}$\\
    FlashVSR-full (frozen)~\cite{flashvsr} & 0.585 & 0.459 & 24.76 & 0.690 & 0.182 & 0.0100 & 0.046 & \best{73.68} & 0.13$^{A}$\\
    RealViFormer~\cite{realviformer} & 0.504 & 0.388 & 24.93 & 0.716 & 0.249 & 0.0077 & 0.058 & 58.97 & 0.02$^{W}$\\
    SeedVR2-3B~\cite{seedvr2} & \second{0.642} & 0.501 & \second{26.08} & \second{0.764} & 0.179 & 0.0078 & 0.044 & 48.00 & 0.49$^{A}$\\
    DicFace$^\dagger$~\cite{dicface} & 0.555 & 0.427 & 25.55 & 0.759 & 0.211 & 0.0068 & 0.073 & 42.66 & 0.30$^{W}$\\
    KEEP$^\dagger$~\cite{keep} & 0.524 & 0.406 & 25.14 & 0.747 & 0.173 & 0.0077 & 0.081 & 55.64 & 0.10$^{W}$\\
    PGTFormer$^\dagger$~\cite{pgtformer} & 0.534 & 0.411 & 25.31 & 0.744 & 0.211 & 0.0070 & 0.060 & 50.35 & 0.31$^{W}$\\
    SVFR$^\dagger$~\cite{svfr} & 0.537 & 0.428 & 25.08 & 0.739 & 0.174 & 0.0063 & 0.026 & 56.29 & 1.48$^{A}$\\
    CodeFormer$^\dagger$~\cite{codeformer} & 0.476 & 0.365 & 24.59 & 0.705 & 0.212 & 0.0099 & 0.126 & 62.74 & 0.07$^{W}$\\
    ReF-LDM$^\dagger$~\cite{refldm} & 0.517 & 0.439 & 23.60 & 0.668 & 0.218 & 0.0212 & 0.338 & 63.85 & 2.29$^{W}$\\
    InstantRestore$^\dagger$ ($K{=}4$)~\cite{instantrestore} & 0.621 & 0.542 & 24.96 & 0.738 & 0.172 & 0.0089 & 0.072 & 57.75 & 0.61$^{A}$\\
    FaceMe$^\dagger$~\cite{faceme} & 0.601 & 0.506 & 25.56 & 0.752 & 0.213 & 0.0095 & 0.156 & 44.42 & 4.94$^{A}$\\
    RGFVR$^\dagger$ (1 ref.)~\cite{rgfvr} & 0.641 & \second{0.559} & 24.32 & 0.743 & \second{0.169} & 0.0079 & 0.032 & 58.29 & 1.35$^{A}$\\
    \midrule
    \method (ours, Ch.1+Ch.2, 20k) & \best{0.670} & \best{0.561} & \best{26.71} & \best{0.794} & \best{0.159} & \best{0.0050} & 0.024 & 52.32 & 0.07$^{W}$/0.04$^{A}$\\
    \bottomrule
    \end{tabular*}
\end{table*}

\vspace{-2mm}
\section{Experiments}
\label{sec:experiments}

\textbf{Data and degradation.}
We construct an identity-disjoint benchmark from VoxCeleb2~\cite{voxceleb2}: 488 identities split into 355/42/91 for training/validation/test. Each identity has an HQ clip at 25 fps (test clips: 111--200 frames) and a session-disjoint album from other videos of the same person. Test albums contain 3--21 unwarped face crops (median 9). The primary heavy track applies two randomized passes of blur, resize, noise, and JPEG following Real-ESRGAN~\cite{realesrgan}, followed by sinc filtering, $4\times$ downsampling, and H.264/H.265 compression at CRF 28--40 on the LR grid. The median face width is 64 pixels in the LQ frame. Training mixes this track with a light track.

\textbf{Training and protocol.}
We train on one A100 with AdamW, learning rate $10^{-4}$, weight decay 0.01, gradient clipping 1.0, one 37-frame clip per step (batch size 1, no gradient accumulation) under a 768-pixel crop cap, and warm-started windows. The primary result uses the final 20k-step checkpoint. For evaluation, we take the first 85 consecutive frames of each clip and append four copies of the last LQ frame for look-ahead, yielding 89 inputs. Only the original 85 frames are scored. HQ target frames are not padded. Metrics are averaged within identity, then equally across identities. Brackets give 95\% percentile intervals from 3,000 bootstrap resamples of identities with replacement (2.5th and 97.5th percentiles of the means). Method differences are formed within identity before resampling. ID-GT is the ArcFace cosine between restored and HQ faces. ID-alb compares restorations with the ArcFace album centroid. All methods use alignment from LQ landmarks. Frames without a valid LQ face detection are omitted from identity metrics. PSNR-F and LPIPS-F use the same $1.3\times$ square face region, resized to $256^2$ for LPIPS-F. Warp is occlusion-weighted error under GT RAFT flow~\cite{raft}. MUSIQ~\cite{musiq} measures no-reference quality. VIDD is mean adjacent-frame ArcFace distance on fixed alignments (GT: 0.031). For crop-based methods, restored $512^2$ faces are inverse-warped and feathered onto bicubic frames under the same landmark protocol ($\dagger$ in Table~\ref{tab:main}).

\textbf{Baselines and runtime.}
The comparison covers bicubic interpolation, generic VSR, blind face image/video restoration, and reference-guided methods. ReF-LDM and FaceMe receive the supported album, InstantRestore receives the four highest-ranked images, and RGFVR receives the highest-quality image. Crop-based methods restore aligned $512^2$ faces before paste-back. Runtime is the median restoration-call time per output frame from preprocessed inputs for every method.  Preprocessing and video writing are excluded for all methods. The restoration ran on an A100 at 0.04 s/frame and on an RTX 4070 Ti at 0.07 s/frame, including per-window UV rendering and patch projection. The frozen backbone has the same rounded median on that card. The measurement excludes offline album encoding and LQ-video DECA fitting. 

\textbf{Model selection.}
We selected the final architecture on 42 validation identities before test evaluation. Validation favored Ch.2 (MUSIQ $+0.90$, $[0.18,1.67]$) and removing the spatial gate (PSNR-F $+0.195$ dB, $[0.147,0.248]$), while Ch.3 showed no clear identity benefit. We then fixed the model and evaluated it on 91 disjoint test identities. Subsequent analyses did not alter the model. 

\vspace{-2mm}
\section{Results and Analysis}
\label{sec:results}

\textbf{Heavy degradation.}
Under heavy degradation, \method improves both identity and fidelity. Differences are paired per-identity means computed on unrounded values and need not equal differences of the rounded entries in Table~\ref{tab:main}. Against the frozen backbone, ID-GT increases by 0.083 ($[0.072,0.095]$), PSNR-F by 1.73 dB, and LPIPS-F decreases by 0.017. Our method improves ID-GT by 0.027 over SeedVR2 ($[0.016, 0.040]$), 0.029 over RGFVR ($[0.014, 0.043]$), and 0.049 over InstantRestore ($[0.032, 0.064]$). \method also improves PSNR-F/LPIPS-F by 0.64 dB/0.020 over SeedVR2 and 2.39 dB/0.010 over RGFVR. RGFVR and \method have similar ID-alb (0.559/0.561). \method's VIDD is 0.024, against 0.049 for frozen FlashVSR, 0.044 for SeedVR2, and 0.031 for the ground truth. 

Frozen FlashVSR has higher MUSIQ (72.3 vs.\ 52.3). On the light track, SeedVR2 leads in ID-GT (0.897/0.887) and LPIPS-F (0.067/0.082), while \method leads in PSNR-F (30.62/29.75\,dB). 

\begin{table}[t]
\centering
\caption{Post hoc heavy-track ablations on 91 test identities. 3-ch. adds geometry Ch.3. ``Global'' denotes global LoRA. $\Delta$ is Ch.1+Ch.2 minus Ch.1-only at 20k, paired per identity using unrounded values.}
\label{tab:abl}
\setlength{\tabcolsep}{1.4pt}
\renewcommand{\arraystretch}{0.94}
\begin{tabular*}{\linewidth}{@{\extracolsep{\fill}}lrrrr}
\toprule
Configuration & ID-GT$\uparrow$ & PSNR-F$\uparrow$ & LPIPS-F$\downarrow$ & MUSIQ$\uparrow$\\
\midrule
3-ch., spatially gated (10k) & 0.667 & 26.51 & 0.163 & 50.72\\
$\quad-$ Ch.1 tokens (10k) & 0.630 & 26.58 & \best{0.157} & 51.53\\
3-ch., global (10k) & \best{0.673} & 26.70 & 0.161 & 51.05\\
Ch.1 only, global (10k) & \best{0.673} & 26.71 & 0.166 & 50.34\\
Ch.1+Ch.2, global (10k) & 0.669 & 26.71 & 0.161 & 51.62\\
\midrule
Ch.1 only, global (20k) & 0.669 & \best{26.73} & 0.159 & 51.63\\
Ch.1+Ch.2, global (20k) & 0.670 & 26.71 & 0.159 & \best{52.32}\\
$\Delta$ Ch.2 at 20k & $+0.001$ & $-0.01$ & $0.000$ & $+0.69$\\
\bottomrule
\end{tabular*}
\end{table}

\begin{figure*}[!t]
\centering
\includegraphics[width=\textwidth]{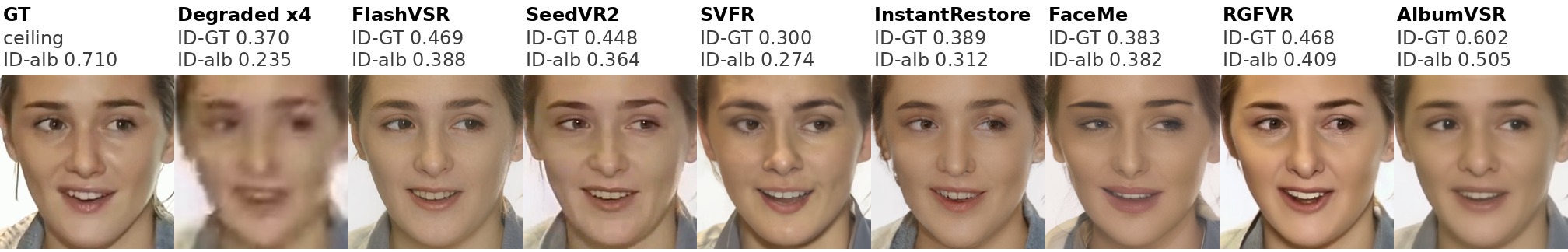}\\[-1.2pt]
\includegraphics[width=\textwidth]{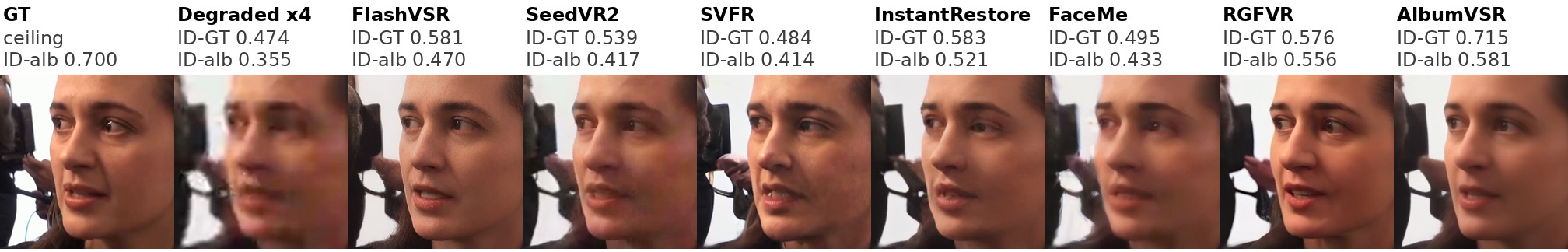}\\[-1.2pt]
\includegraphics[width=\textwidth]{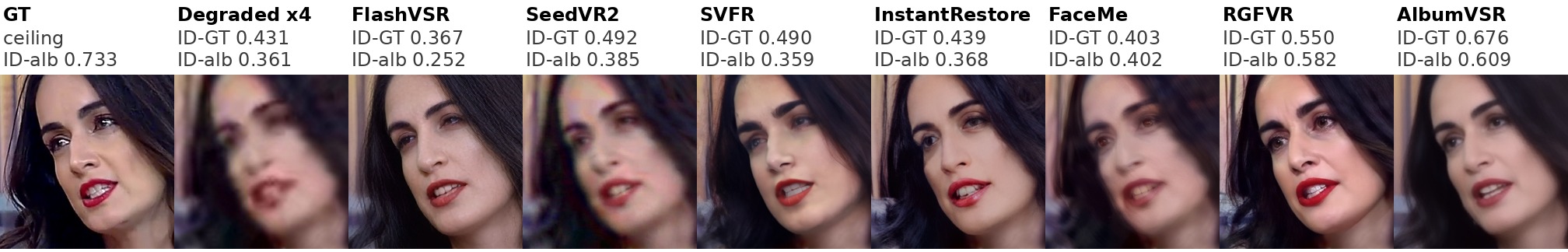}\\[-1.2pt]
\includegraphics[width=\textwidth]{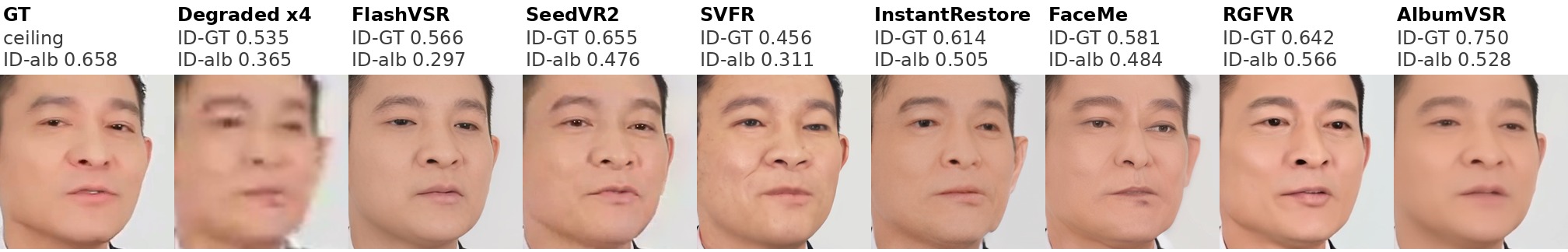}

\vspace{2pt}

\includegraphics[width=0.544\textwidth]{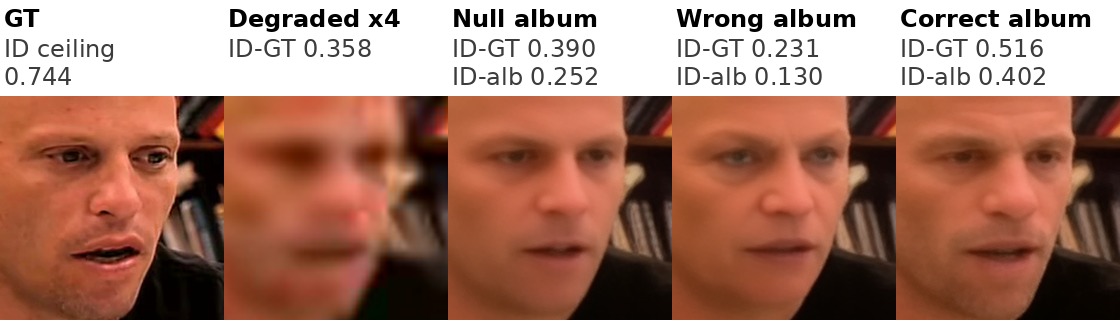}\hfill
\includegraphics[width=0.436\textwidth]{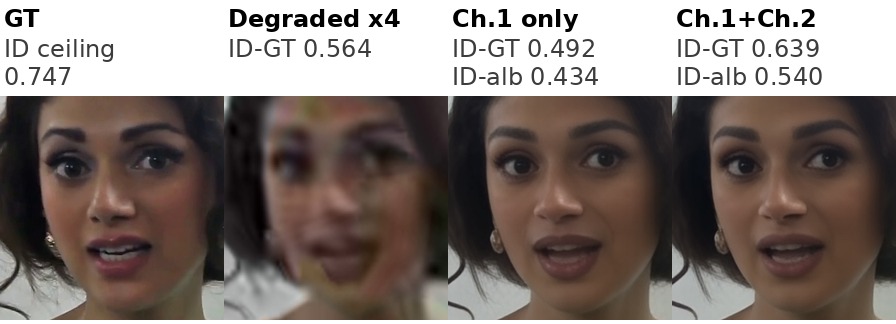}

\caption{\textbf{Top:} Four heavy-degradation examples: id00068/047, id00109/013, id00456/050, and id00560/010. We show GT, degraded input, FlashVSR, SeedVR2, SVFR, InstantRestore, FaceMe, RGFVR, and \method. Labels report per-frame ArcFace ID-GT/ID-alb. The GT ID-alb values show the similarity between the target frames and the album. Table~\ref{tab:main} reports all 14 baselines on 91 test identities. \textbf{Bottom:} Qualitative ablations. Left: null, wrong, and correct albums using a separate 14k three-channel model. Right: Ch.1-only versus Ch.1+Ch.2 at 20k. Values are per-frame scores. Table~\ref{tab:abl} reports aggregate results.}
\label{fig:qual_ablation}
\end{figure*}


\textbf{Ablations.}
Table~\ref{tab:abl} analyzes the final two-channel model against earlier configurations. Ch.3 adds two album-shape tokens and per-frame LQ expression/pose tokens. Removing Ch.1 lowers ID-GT from 0.667 to 0.630 but also lowers LPIPS-F from 0.163 to 0.157. Removing Ch.3 changes ID-GT by $-0.004$ and MUSIQ by $+0.57$. At 20k, Ch.2 changes ID-GT by $+0.001$ ($[-0.004,0.006]$) and PSNR-F by $-0.01$ dB, but raises MUSIQ by 0.69 ($[0.24,1.17]$).

\textbf{Album interventions.}
With a separate 14k three-channel model, a wrong-identity album lowers ID-GT by 0.088 ($[-0.103,-0.074]$) and ID-alb by 0.108, while full-frame PSNR changes by only $-0.03$ dB. Zeroing album tokens lowers ID-GT by 0.054. Frozen FlashVSR, zero-album, and correct-album models score 0.586, 0.607, and 0.661 ID-GT, respectively.

Restricting Ch.1 to the first valid album image lowers ID-GT from 0.661 to 0.657 ($-0.004$, $[-0.006,-0.002]$). Ch.2 and pooled shape still use the full album, so this isolates multi-reference aggregation in Ch.1.

\textbf{Efficiency and state reuse.}
A window adds 68 tokens (four from Ch.1 and 64 from Ch.2) to each DiT block. If no face-valid render exists, the model uses Ch.1 alone. The prompt K/V remain cached and only the dynamic album K/V are formed for the current window. The model has approximately 23M trainable parameters: 8.85M in LoRA, 5.65M in the set encoder, 7.96M in the patch projection, and 0.39M in the temporal tags, corresponding to about 1.3\% of the 1.75B frozen backbone. The first DiT evaluation processes a 21-frame window. Each subsequent evaluation advances the stream by eight frames. On the RTX 4070 Ti, both \method and the frozen backbone have a rounded median of 0.07 s/frame.

\textbf{Identity distribution.}
AlbumVSR has higher ID-GT than frozen FlashVSR for 96\% of the 91 identities. Improvements over SeedVR2 and RGFVR are more heterogeneous, occurring for 62\% and 74\% of identities, respectively.

\textbf{Limitations.}
Several limitations remain. First, the benchmark comes from one source dataset and uses synthetic degradation. Real camera, surveillance, and archival artifacts have not been tested. Second, the albums are curated, session-disjoint, and relatively small. Robustness to weak or contaminated albums remains unknown. Third, Table~\ref{tab:main} excludes offline atlas construction and the per-video DECA fit on degraded frames. The reported 0.07 s/frame is therefore not an end-to-end live-streaming latency. 

\vspace{-2mm}
\section{Conclusion}
\label{sec:conclusion}
On heavily degraded clips, album conditioning raises ID-GT from 0.586 to 0.670 and face PSNR from 24.98 to 26.71 dB relative to frozen FlashVSR. \method injects four identity tokens and view-aligned texture context into a frozen one-step streaming VSR model, with global LoRA adapting latent-side attention. It improves identity and face fidelity over generic, face-specific, and reference-guided baselines while matching the frozen backbone's rounded runtime. On the light track, SeedVR2 remains better in ID-GT and LPIPS-F.

\vspace{-2mm}
\section{Compliance with Ethical Standards}
This study analyzed previously released VoxCeleb2 videos~\cite{voxceleb2}. No participants were recruited or recorded for this work.

\vspace{-2mm}
\section{Acknowledgments}
This work was partly supported by a Purdue University research assistantship. The authors declare no known competing interests.

\vspace{-2mm}
\bibliographystyle{IEEEbib}
\bibliography{refs}

\end{document}